\documentclass{OAGM}
\OAGMarXiv{1304.1876}

\usepackage{graphicx}

\title{Reading Ancient Coin Legends: Object Recognition vs. OCR }

\author{Albert Kavelar, Sebastian Zambanini \and Martin Kampel\\
  Computer Vision Lab, Vienna University of Technology, Austria}

\begin{document}
\maketitle

\begin{abstract}
Standard OCR is a well-researched topic of computer vision and can be considered solved for machine-printed text. However, when applied to unconstrained images, the recognition rates drop drastically. Therefore, the employment of object recognition-based techniques has become state of the art in scene text recognition applications. This paper presents a scene text recognition method tailored to ancient coin legends and compares the results achieved in character and word recognition experiments to a standard OCR engine. The conducted experiments show that the proposed method outperforms the standard OCR engine on a set of 180 cropped coin legend words.
\end{abstract}

% ====================
% === INTRODUCTION ===
% ====================

\section{Introduction}\label{sec:Introduction}
OCR is a well-researched subject in computer vision \cite{Plamondon2000,Suen1980,Vinciarelli2002}. A classical off-line OCR system \cite{Plamondon2000,Vinciarelli2002} comprises the following four steps: preprocessing, normalization, segmentation and detection. The first step, preprocessing, incorporates background and noise removal and binarization. The latter is performed using either a global threshold such as Otsu's method \cite{Otsu1979} or a locally adaptive threshold like Sauvola's approach \cite{Sauvola2000}. Both techniques depend on a rather bimodal gray value distribution (either global or in a local pixel neighborhood), which allows separating the text from the background. This approach works well for texts written on homogeneous backgrounds such as a sheet of paper or road signs, where the text color contrasts strongly with the background color for an optimal legibility. When it comes to coin legends where the textual inscription, the so-called \textit{legend}, is simply embossed in the metal, and no different color or alloy is used, binarization of classical OCR methods \cite{Plamondon2000,Vinciarelli2002} becomes error-prone because text and background color are identical; the only visible information is the highlights and shadows resulting from the coin's relief surface structure.

\begin{figure}[t]
 \centering
 \includegraphics[width=0.6\textwidth]{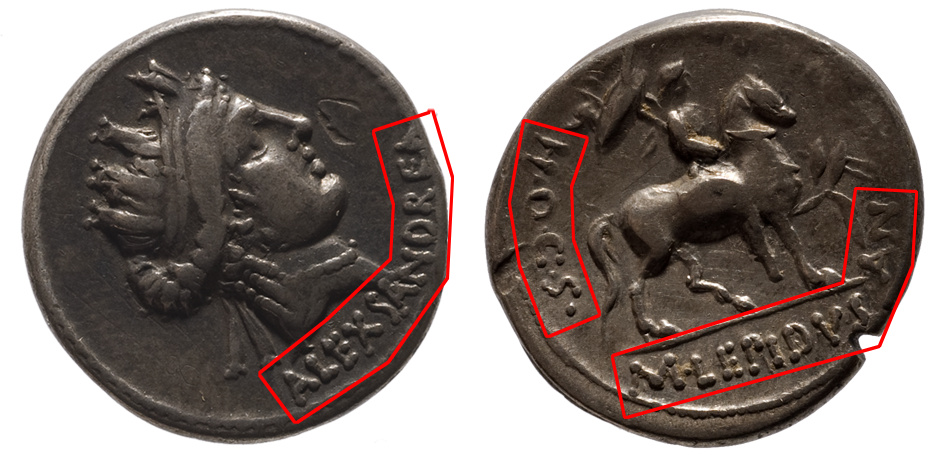}
 \caption{Obverse and reverse of an ancient Roman Republican coin with the legend highlighted in red.}
 \label{fig:introCoinExample}
\end{figure}

Even in case of a successful binarization, traditional OCR methods would have difficulties performing normalization because skew compensation methods are designed for certain document layouts and rely on text regions having a prevailing text orientation. Legends of ancient coins only comprise a few words which cover a small part of the coin (see Fig.~\ref{fig:introCoinExample}). However, without horizontal text alignment, standard OCR engines such as the ABBYY FineReader\footnote{http://finereader.abbyy.com/} are not able identify letters correctly. Thus, methods for recognizing texts in unconstrained images must rely on binarization-free methods and rather follow object recognition techniques than the traditional OCR pipeline. Wang et al. \cite{Wang2011} state that recognizing text in unconstrained images can be broken down into four subproblems: (1) cropped character classification, (2) full image text detection, (3) cropped word recognition and (4) full image word recognition. Kavelar et al. have shown that Wang's \textit{scene text recognition} (STR) pipeline can be enhanced in a way that it can handle arbitrary text orientations and can be used for recognizing ancient coin legends \cite{Kavelar2012}. However, they did not compare their method to standard OCR software. As stated above, standard OCR has difficulties detecting and normalizing text regions in images containing little amounts of text. Thus, this work is focused on cropped character and word recognition rather than on full image word recognition, to allow for a comparison between the presented method and a standard OCR engine.

The remainder of this paper is organized as follows: Section \ref{sec:RelatedWork} reviews the state of the art in \textit{scene text recognition} (STR) and object recognition-based character recognition. Our word recognition system is described in Section \ref{sec:Methodology}. Section \ref{sec:Experiments} evaluates the proposed methodology and a standard OCR engine on a set of cropped characters and legend words. Finally, Section \ref{sec:Conclusion} concludes this paper and draws an outlook for further research.

% ====================
% === RELATED WORK ===
% ====================

\section{Related Work}\label{sec:RelatedWork}
De Campos et al.\cite{deCampos2009} propose a method towards reading text in images of unconstrained scenes. In the context of STR, additional problems need to be considered: geometric distortion resulting from camera positions, unconstrained illumination conditions, arbitrary image resolutions and a wide range of font families and styles \cite{deCampos2009}. They describe STR as a multi-stage process comprising the following steps: (1) text localization; (2) character and word segmentation; (3) character and word recognition and (4) inclusion of language models and context. They focus on the character recognition aspect of STR to prove the feasibility of adopting object recognition techniques. Characters are described in a bag-of-visual-words representation. Various local image descriptors and classifier combinations are benchmarked and show that object recognition is an adequate approach towards character recognition in unconstrained images problems.

Wang and Belongie \cite{Wang2010} took STR one step further by working on cropped word images rather than on cropped characters. They employ \textsc{Hog} features to describe individual characters, which are then classified using a nearest neighbor classifier, since this combination gave the best results in their experiments and even outperforms the method proposed by de Campos et al. \cite{deCampos2009}. After applying character segmentation to the cropped word image, each character is normalized to match a standard height and aspect ratio. Next, \textsc{Hog} features are computed for each character, which can be compared via normalized cross-correlation (NCC). When scanning an input image for characters, all training images (i.e., all templates) for each character class are resized to the height of the input image and shifted to every possible location where the \textsc{Hog} features of the template are compared to the ones of the underlying image location using NCC and the highest value is selected for every class. Finally, the algorithm searches for every word of a given lexicon using \textit{pictorial structures} \cite{Felzenszwalb2005}. The word causing the lowest costs in the pictorial structures model is detected.

In \cite{Wang2011}, Wang et al. further extended this method to a fully fledged STR algorithm which covers the entire STR pipeline described by de Campos et al. \cite{deCampos2009}. As in their previous approach, \textsc{Hog} features are used to represent characters. However, instead of NCC, random ferns \cite{Bosch2007} are used for assigning scores to candidate character locations. The remainder of the algorithm closely follows the method described in \cite{Wang2010}. The main finding of this work is that STR does not perform significantly worse when the initial text recognition step is omitted.

% ===================
% === METHODOLOGY ===
% ===================

\section{Methodology}\label{sec:Methodology}
\begin{figure}[t]
	\centering
	\includegraphics[width=1.0\textwidth]{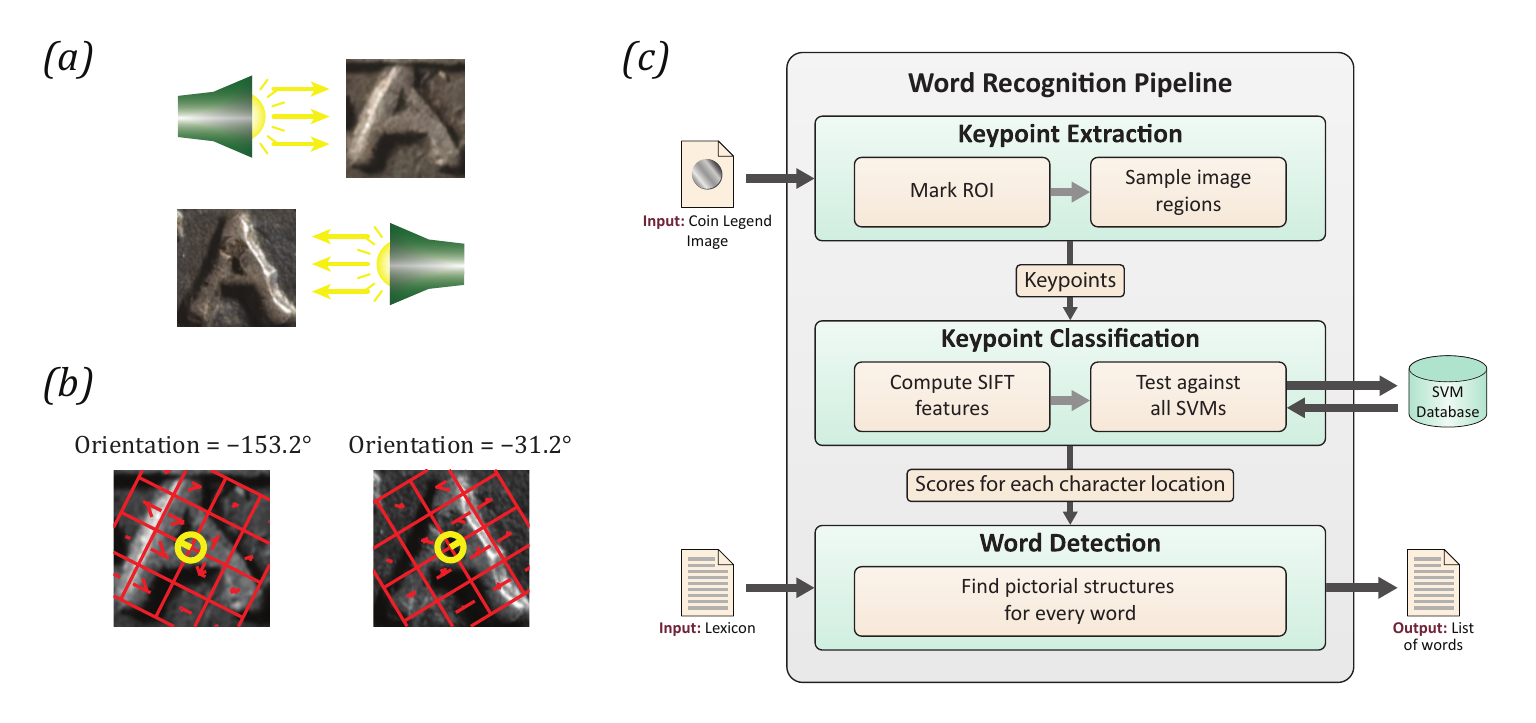}
	\caption{(a) The coin's relief structure causes letters to appear different depending on the light's angle of incidence. (b) Different \textsc{Sift} descriptor orientations resulting from different light source directions. (c) Word recognition pipeline.}
	\label{fig:methCombined}
\end{figure}

Our approach is based on the ideas of Wang et al. \cite{Wang2011}. Instead of the \textsc{Hog} features proposed by Wang et al., an adapted version of \textsc{Sift} features, which respects the relief structure of coin surfaces and only uses half the angular spectrum to compensate for light sources illuminating the coin from opposite directions, is employed. Mapping gradients of opposite directions to the same orientation bin of the \textsc{Sift} descriptor takes into account that illuminating letters from opposite directions casts shadows in opposite directions (see Fig.~\ref{fig:methCombined}(a)) and therefore results in different \textsc{Sift} descriptors, as shown in Fig.~\ref{fig:methCombined}(b). Using only half the angular spectrum alleviates this problem, since it reduces the number of possible \textsc{Sift} descriptors for a legend letter and thus increases the chances of a correct recognition. As opposed to Kavelar et al. \cite{Kavelar2012} who tested legend recognition on entire coin images, this work focuses on images of cropped letters and words. The general architecture of the proposed system is illustrated in Fig.~\ref{fig:methCombined}(c).

%\begin{figure}[t]
%	\centering
%	\includegraphics[width=0.8\textwidth]{./figures/methLegendRecognitionPipeline2.pdf}
%	\caption{Word Detection Pipeline.}
%	\label{fig:methPipeline}
%\end{figure}

\subsection{SVM Training}\label{meth:Training}
In the first step, the character appearances are learned and a support vector machine (SVM) is trained. In order to describe a character, a single centered \textsc{Sift} descriptor which spans the entire character is used. \textsc{Sift} offers rotational invariance, and while in other scenarios this is desirable, we sacrifice this additional degree of freedom for a gain in classification performance (see Section \ref{sec:Experiments}). That is, the orientation of the \textsc{Sift} descriptor is constrained to be aligned horizontally. In the $35$ legend words considered, $19$ different letters occur. However, the letter 'I' is not considered because it is contained in many other letters (such as 'H' or 'T'), which gives an overall of $18$ different character classes. 
%The training set comprises $50$ manually segmented legend letter images of a standardized size of $100 \times 100$ for each class and the optimal kernel parameters are determined via 5-fold cross validation.

\subsection{Word Recognition}\label{meth:Recognition}
The word recognition pipeline is depicted in Fig.~\ref{fig:methCombined}(c). In the \textit{keypoint extraction step}, the region-of-interest (ROI) is marked, which - due to the known layout of the cropped legend word images - simply is a rectangular area having a spacing of a quarter of the image height to each border. The spacing was chosen to allow for slight variations in character height and placement within the legend word. This ROI is densely sampled to create a grid of candidate character locations, which are passed to the \textit{keypoint classification step}. In this step, a horizontally aligned \textsc{Sift} descriptor is computed for each candidate character location. The scale of the descriptor was chosen manually based on the layout of the input images and was set to $\frac{3 \cdot H}{4}$, where $H$ is the image height. Next, every \textsc{Sift} descriptor is tested against the SVMs trained initially and receives a score for each class that indicates how likely it is to encounter the respective letter at this image location. That is, this step outputs a list of character scores for each pixel in the ROI grid, which is passed to the final \textit{word detection step}. This step measures how close the character configuration of a certain lexicon word can be matched to the image. This is accomplished using pictorial structures \cite{Fischler1973}, which were rediscovered for object recognition by Felzenszwalb and Huttenlocher \cite{Felzenszwalb2005}. A pictorial structure model can be thought of as a mass-spring model describing the ideal configuration of an object comprising multiple sub-parts \cite{Fischler1973}. In its optimal configuration, where all sub-parts are arranged in the desired relative distances, no tension is applied to the springs. Translating this model to object recognition means that two aspects need to be considered: (1) How close the model's sub-parts match the underlying image location, that is, the \textit{matching costs} \cite{Felzenszwalb2005}; (2) How heavily the matching deforms the model, i.e., how much tension needs to be applied to the springs, referred to as \textit{deformation costs} \cite{Felzenszwalb2005}. In the context of word detection, a word is considered as an object and its letters are the associated sub-parts. In order to search for a specific word, the algorithm tries placing the letters at every possible candidate character location, thus evaluating each possible word configuration. To narrow down the set of configurations, the following restrictions are imposed: (1) Two letters must not intersect. (2) The distance between two consecutive characters must not exceed a threshold $\theta$. (3) Words are assumed to run from left to right.

Among all configurations, the one causing the lowest costs is chosen as the optimal configuration for this word, and the word having the lowest costs is detected. Mathematically, an optimal word configuration can be found when the problem is formulated as a weighted graph optimization problem. Let $\mathcal{K} = \{k_1, \ldots, k_m\}$ be the set of the $m$ possible candidate character locations found in the keypoint localization step. The subset $\mathcal{L} = \{l_1, \ldots, l_n\} \subseteq \mathcal{K}$ is the set of character locations of a certain $n$ letters long word configuration. The multi set of letters of the respective word is given by \mbox{$C = \{c_1, \ldots, c_p | c_i \in \mathcal{A}\}$}, where $\mathcal{A}$ is the used alphabet. The directed graph representing the word configuration is given by $G(C,E)$, where $C$ is the list of the $n$ characters $c_i$ located at $l_i \in \mathcal{L}$ and $E = \{e_j, \ldots, e_{n-1}\}$ is the list of the $n-1$ edges $e_j(l_j, j_{j+1})$ connecting adjacent characters. Those edges can be thought of as the pictorial structure's conceptual springs \cite{Wang2010} mentioned above. In order to find the optimal word configuration, the objective function

\begin{equation}
	\mathcal{L^{*}} = \min_{\forall i, l_i \in \mathcal{L}}(\lambda\sum\limits_{i=1}^{n}{s_i(l_i, c_i)} + (1-\lambda)\sum\limits_{i=1}^{n-1}d(l_i, l_{i+1}))
\label{eq:pictorialStructureObjectiveFunction}
\end{equation}

has to be minimized where $\mathcal{L^{*}} = \{l_1^{*}, \ldots, l_n^{*}|l_i^{*} \in \mathcal{L}\}$ expresses a specific configuration within the image \cite{Wang2011}. This means, the lower the score achieved, the better the word is recognized in the image. In Eq. \ref{eq:pictorialStructureObjectiveFunction}, $\lambda$ is a trade-off parameter which allows balancing the contribution of matching and deformation costs and has to be determined empirically \cite{Wang2011}. Recasting Eq. \ref{eq:pictorialStructureObjectiveFunction} to a recursive function allows solving the optimization problem by using dynamic programming \cite{Wang2011}. The reformulation leads to 

\begin{equation}
	D(l_i) = \lambda s(l_i, c_i) + (1 - \lambda) \min_{l_{i+1} \in \mathcal{L}}{d(l_i, l_{i+1})} + D(l_{i+1}),
\label{eq:pictorialStructureRecursiveFunction}
\end{equation} 

where the position of the $i$-th character is fixed at the location $l_i$. Thus, the costs of the optimal configuration are given by $\min_{l_1 \in \mathcal{L}}D(l_1)$ \cite{Wang2011}. To guarantee a fair comparison between words of different lengths, the resulting score is divided by the number of letters contained in the respective word. After Eq. \ref{eq:pictorialStructureRecursiveFunction} has been solved for every lexicon word, the one resulting in the lowest costs is considered detected.

% ===================
% === EXPERIMENTS ===
% ===================

\section{Experiments}\label{sec:Experiments}
\begin{figure}[t]
	\centering
	\includegraphics[width=1.0\textwidth]{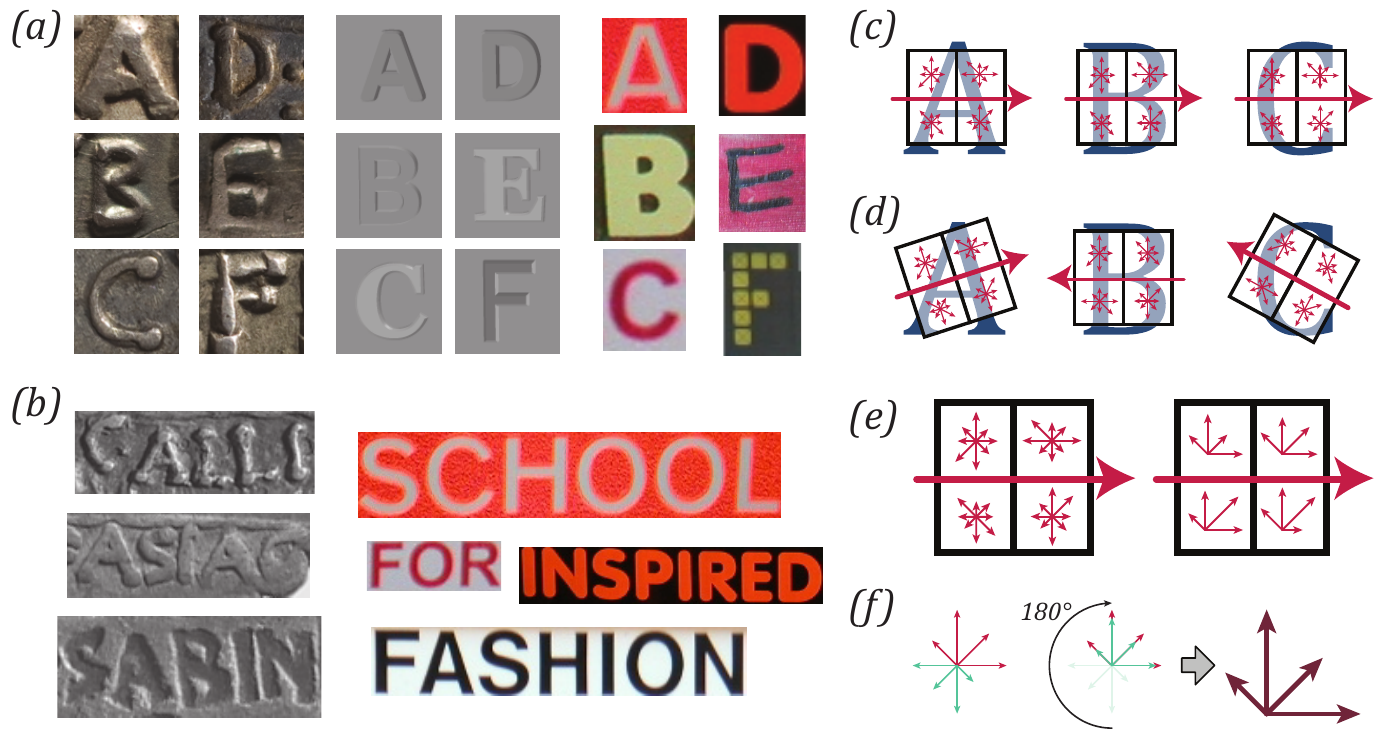}
	\caption{(a) Samples of the character recognition test set. From left to right: \textit{Coin}, \textit{Synth}, \textit{ICDAR}. (b) Samples of the word recognition test set. (c) Fixed descriptor orientations. (d) Dynamic descriptor orientations based on the dominant gradient direction. (e) Full vs. half angular spectrum. (f) Constructing the half-spectrum by adding up magnitudes of inverted gradient directions.}
	\label{fig:expCombined}
\end{figure}

This section presents the results the proposed method and the standard OCR engine ABBYY achieved in the cropped word and character recognition experiments carried out on test sets for manually segmented legend letters and words (\textit{Coin}), cropped letter images created synthetically using a standard vector graphics editor, which mimic the appearance of legend letters (\textit{Synth}) as well as cropped letter and word images of the ICDAR 2003 dataset (\textit{ICDAR}). Fig.~\ref{fig:expCombined}(a) and \ref{fig:expCombined}(b) give examples of the datasets used. The \textit{Coin} training set consists of $50$ $100 \times 100$ pixel sized images for $18$ classes, i.e., $900$ images; the test set comprises $5$ images per $18$ classes giving an overall of $90$ images. The \textit{Synth} training set contains $50$ images for $18$ classes and the test set comprises $10$ images for the same $18$ classes. From the ICDAR 2003 cropped character dataset, a subset for the same $18$ character classes was selected as a test set containing a total of $156$ images. The respective training set consists of $932$ images; up to $60$ images per class are used, depending on how many images the ICDAR 2003 dataset provides for this letter. The cropped legend word dataset comprises $180$ images and the subset chosen from the ICDAR 2003 cropped word recognition set comprises $95$ images showing words that can be written with the $18$ characters of the alphabet used. The alphabet of the ABBYY reader was configured to comprise the same $18$ characters which the SVMs were trained for; and in the word recognition experiments the same $35$-word lexicon as in the tests with the proposed algorithm was used.

\subsection{Cropped Character Recognition}\label{subSec:CroppedCharacterRecognition}
The character recognition performance of the \textsc{Sift} descriptor was tested with two different configurations: The first setting uses the entire angular range from $[0^{\circ}, \cdots, 360^{\circ})$ for gradient directions, whereas the second configuration only uses half the angular range, i.e., $[0^{\circ}, \cdots, 180^{\circ})$. Fig.~\ref{fig:expCombined}(c) - (f) illustrate the difference between \textsc{Sift} descriptors having fixed (Fig.~\ref{fig:expCombined}(c)) and dynamic orientations (Fig.~\ref{fig:expCombined}(d)) as well as the difference between full and half angular spectrum (Fig.~\ref{fig:expCombined}(e)) and how half the spectrum is constructed from the full spectrum (Fig.~\ref{fig:expCombined}(f)).

%\begin{figure}[t]
%	\centering
%	\includegraphics[width=0.8\textwidth]{./figures/expSiftDescriptorOptions.pdf}
%	\caption{Various \textsc{Sift} descriptor configurations. (a) Fixed descriptor orientations. (b) Dynamic descriptor orientations based on the dominant gradient direction. (c) Full vs. half angular spectrum. (d) Constructing the half-spectrum by adding up magnitudes of diametrically opposed gradient directions.}
%	\label{fig:expSiftConfigurations}
%\end{figure}

Besides the aforementioned \textsc{Sift} configurations, the SVMs were once trained with an additional background class comprising randomly chosen background snippets containing no legend letters. The results for the cropped character recognition are listed in Tab.~\ref{tbl:characterRecognitionAccuracy}. The optimal parameters for the SVMs used in the character recognition process were found using $5$-fold cross-validation. 

%	  								& $Coin$ & $ICDAR$ \\
%	\hline
%	Proposed method 	& 37.8\% & 48.4\% \\
%	\hline
%	ABBYY							& --- 	 & 57.9\% \\

\begin{table}[!t]
\renewcommand{\arraystretch}{1.1}
\caption{Character and Word Recognition \textit{(W)} Accuracy}
\label{tbl:characterRecognitionAccuracy}
\centering
	\begin{tabular}{|c||c|c|c||c|c|}
	\hline
	  										& \textit{Coin} 	 & \textit{Synth}  & \textit{ICDAR}    	& \textit{Coin} (W) 	& \textit{ICDAR} (W) \\
	\hline
	$360^{\circ}$, no BG 	& 64.4\% & 78.9\% & 72.3\% 		& --- 				& --- \\
	\hline
	$180^{\circ}$, no BG 	& 75.6\% & 83.9\% & 72.5\% 		& 37.8\% 			& 48.4\% \\
	\hline	
	$360^{\circ}$, BG 		& 68.9\% & --- 		& --- 			& --- 				& --- \\
	\hline
	$180^{\circ}$, BG 		& 72.2\% & --- 		& --- 			& --- 				& --- \\
	\hline	
	ABBYY 								& --- 	 & 20.6\% & 46.8\% 		& --- 				& 57.9\% \\
	\hline
	\end{tabular}
\end{table}

\begin{table}[!t]
	\renewcommand{\arraystretch}{1.1}
	\caption{False Negative (FN) and False Positive (FP) Rates per Class for the \textit{Coin} Dataset}
	\label{tbl:falseNegativeRates}
	\centering
		\begin{tabular}{|l||cc|cc|cc|cc|}
			\hline
					\rotatebox{90}{Class} & \rotatebox{90}{180, no BG, FN } & \rotatebox{90}{180, no BG, FP} & \rotatebox{90}{360, no BG, FN } & \rotatebox{90}{360 no BG, FP } & \rotatebox{90}{180, BG, FN } & \rotatebox{90}{180, BG, FP} & \rotatebox{90}{360, BG, FN } & \rotatebox{90}{360 BG, FP } \\

					\hline
						a & $0.20\%$ & $0.00\%$ & $0.20\%$ & $0.00\%$ & $0.20\%$ & $0.00\%$ & $0.20\%$ & $0.00\%$ \\
						b & $0.80\%$ & $0.02\%$ & $1.00\%$ & $0.02\%$ & $0.60\%$ & $0.04\%$ & $1.00\%$ & $0.01\%$ \\
						c & $0.20\%$ & $0.04\%$ & $0.40\%$ & $0.02\%$ & $0.20\%$ & $0.02\%$ & $0.40\%$ & $0.02\%$ \\
						d & $0.20\%$ & $0.01\%$ & $0.60\%$ & $0.00\%$ & $0.00\%$ & $0.01\%$ & $0.00\%$ & $0.01\%$ \\
						e & $0.00\%$ & $0.00\%$ & $0.40\%$ & $0.02\%$ & $0.00\%$ & $0.00\%$ & $0.40\%$ & $0.00\%$ \\
						f & $0.20\%$ & $0.01\%$ & $0.00\%$ & $0.00\%$ & $0.00\%$ & $0.02\%$ & $0.00\%$ & $0.00\%$ \\
						g & $1.00\%$ & $0.02\%$ & $0.80\%$ & $0.02\%$ & $0.80\%$ & $0.02\%$ & $0.60\%$ & $0.02\%$ \\
						h & $0.00\%$ & $0.00\%$ & $0.00\%$ & $0.00\%$ & $0.00\%$ & $0.00\%$ & $0.00\%$ & $0.00\%$ \\
						l & $0.20\%$ & $0.00\%$ & $0.20\%$ & $0.00\%$ & $0.20\%$ & $0.00\%$ & $0.00\%$ & $0.00\%$ \\
						m & $0.00\%$ & $0.00\%$ & $0.00\%$ & $0.00\%$ & $0.20\%$ & $0.01\%$ & $0.00\%$ & $0.00\%$ \\
						n & $0.40\%$ & $0.00\%$ & $0.60\%$ & $0.00\%$ & $0.60\%$ & $0.00\%$ & $0.80\%$ & $0.00\%$ \\
						o & $0.40\%$ & $0.00\%$ & $0.40\%$ & $0.01\%$ & $0.60\%$ & $0.01\%$ & $0.40\%$ & $0.00\%$ \\
						p & $0.20\%$ & $0.01\%$ & $0.20\%$ & $0.06\%$ & $0.20\%$ & $0.02\%$ & $0.20\%$ & $0.02\%$ \\
						r & $0.20\%$ & $0.02\%$ & $1.00\%$ & $0.00\%$ & $0.40\%$ & $0.02\%$ & $0.80\%$ & $0.01\%$ \\
						s & $0.00\%$ & $0.01\%$ & $0.00\%$ & $0.02\%$ & $0.00\%$ & $0.01\%$ & $0.00\%$ & $0.02\%$ \\
						t & $0.00\%$ & $0.00\%$ & $0.20\%$ & $0.02\%$ & $0.00\%$ & $0.00\%$ & $0.20\%$ & $0.01\%$ \\
						v & $0.00\%$ & $0.00\%$ & $0.00\%$ & $0.00\%$ & $0.00\%$ & $0.00\%$ & $0.00\%$ & $0.00\%$ \\
						x & $0.40\%$ & $0.00\%$ & $0.40\%$ & $0.01\%$ & $0.60\%$ & $0.00\%$ & $0.60\%$ & $0.01\%$ \\		
					\hline
					Total & $0.24\%$ & $0.01\%$ & $0.36\%$ & $0.01\%$ & $0.27\%$ & $0.01\%$ & $0.31\%$ & $0.01\%$ \\		
					Abs & $(22)$ & $(13)$ & $(32)$ & $(19)$ & $(26)$ & $(19)$ & $(29)$ & $(13)$ \\		
					\hline
		\end{tabular}
\end{table}

As shown in Tab.~\ref{tbl:characterRecognitionAccuracy}, the proposed algorithm outperforms the standard OCR engine on all three datasets, reaching a recognition rate of $75.6\%$ for the cropped coin legend letters dataset when fixed orientations are used in combination with the half angular range and when the SVMs are trained without the additional background class. The ABBYY reader is capable of recognizing characters of the \textit{Synth} and \textit{ICDAR} dataset using the built-in patterns for character recognition. When applied to the \textit{Coin} dataset, initially no images are detected correctly because of binarization errors, which impede the localization of text regions. However, ABBYY offers to train user patterns instead of using the built-in patterns. Nevertheless, ABBYY fails to correctly detect connected components covering entire letters for nearly all images of the training set (see Fig.~\ref{fig:expResults}(a)) and thus cannot be trained properly. As a consequence, ABBYY fails on the \textit{Coins} dataset. ABBYY has difficulties detecting text regions in low contrast images in general, which explains the lower accuracy achieved on the \textit{Synth} dataset compared to the \textit{ICDAR} dataset. The results on the \textit{Coin} dataset are better for the \textsc{Sift}180 configuration when no background images are used. In case of \textsc{Sift}360, the use of background images slightly increases the overall classification performance. This results from the fact that the \textsc{Sift}360 descriptor provides a richer description of the respective image patch and therefore allows to train the SVMs more accurately. This leads to fewer false positives and false negatives, as shown in Tab.~\ref{tbl:falseNegativeRates}.

\subsection{Cropped Word Recognition}\label{subSec:CroppedWordRecognition}
\begin{figure}[t]
	\centering
	\includegraphics[width=1.0\textwidth]{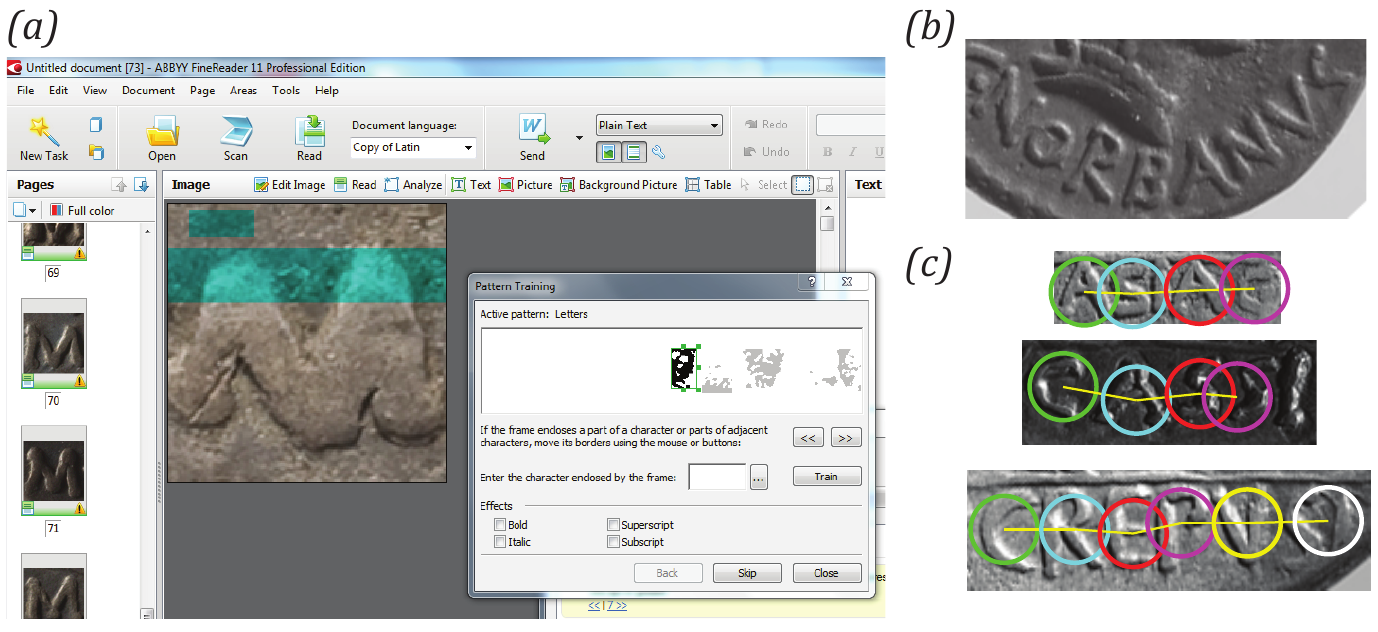}
	\caption{(a) Binarization errors when training the ABBYY reader for coin legend letters. (b) Curved legends with misaligned letters. (c) Correctly detected legend words (from top to bottom: \textit{ASIAG}, \textit{CASSI}, \textit{CREPVSI})}
	\label{fig:expResults}
\end{figure}

The cropped word recognition for the \textit{Coin} dataset was carried out using the \textsc{Sift} descriptor configuration which performed best in the character recognition experiments, i.e., fixed orientation, half-spectrum, trained without background images. For the \textit{ICDAR} dataset, fixed orientations and the use of the full angular spectrum achieved the best accuracy in the character recognition experiments; hence, this setting was used for word recognition. The classification accuracy of the two methods is listed in the last two columns of Tab.~\ref{tbl:characterRecognitionAccuracy}. Even though the character recognition results for the two datasets are similar, the proposed word recognition method works significantly better on the \textit{ICDAR} than on the \textit{Coin} dataset. This results from the fact that for curved legends, even when cropped, not all letters are horizontally aligned (see Fig.~\ref{fig:expResults}(b)). In such a case, not only the horizontal \textsc{Sift} descriptor works worse but also its scale mismatches the letters, since is automatically selected based on the image height. Fig.~\ref{fig:expResults}(c) shows three examples that were recognized correctly. While the ABBYY reader fails again to detect words in the \textit{Coin} images, it achieves a better result on the \textit{ICDAR} dataset, because many of its images are cropped words of traffic signs or grocery store signs (see Fig.~\ref{fig:expCombined}(b)), which still provide all the properties of printed text: sharp contours, high contrast between fore- and background, consistent character spacing and size.

%\begin{table}[!t]
%\renewcommand{\arraystretch}{1.3}
%\caption{Word Recognition Accuracy}
%\label{tbl:wordRecognitionAccuracy}
%\centering
%	\begin{tabular}{|c||c|c|}
%	\hline
%	  								& $Coin$ & $ICDAR$ \\
%	\hline
%	Proposed method 	& 37.8\% & 48.4\% \\
%	\hline
%	ABBYY							& --- 	 & 57.9\% \\
%	\hline	
%	\end{tabular}
%\end{table}

% ==================
% === CONCLUSION ===
% ==================

\section{Conclusion}\label{sec:Conclusion}
This work shows that standard OCR engines are inappropriate for recognizing coin legends since they rely on binarization. Even the use of built-in training mechanisms cannot circumvent this limitation because the connected components detected in the binarization step hardly ever coincide with entire characters. The presented binarization-free technique using tailored \textsc{Sift} descriptors, respects the challenges introduced by the coins relief surface and achieves promising recognition rates on a set of 180 images of cropped coin legends. Future research will explore how multi-view integration affects the legend recognition performance for a subset of images. Furthermore, additional local image features will be evaluated.

\section*{Acknowledgments}
This research is funded by the Austrian Science Fund (FWF): $TRP140-N23-2010$.
%
%This work was supported by whomever. We thank Prof.~X for her
%thoughtful comments.

% \bibliography{refs}
\bibliography{oagmBibliography}

% that's all folks
\end{document}